\documentclass{IEEEtran}
\usepackage{graphicx, amsmath, fancyhdr}
\begin{document}

\title{Clustering of illustrations by atmosphere \\using a combination of \\supervised and unsupervised learning}

\author{\IEEEauthorblockN{Keisuke Kubota, Masahiro Okuda}
\IEEEauthorblockA{\\Doshisha University}}

\maketitle
\thispagestyle{empty}

\begin{abstract}
The distribution of illustrations on social media, such as Twitter and Pixiv has increased with the growing popularity of animation, games, and animated movies. The "atmosphere" of illustrations plays an important role in user preferences. Classifying illustrations by atmosphere can be helpful for recommendations and searches. However, assigning clear labels to the elusive "atmosphere" and conventional supervised classification is not always practical. Furthermore, even images with similar colors, edges, and low-level features may not have similar atmospheres, making classification based on low-level features challenging.
In this paper, this problem is solved using both supervised and unsupervised learning with pseudo-labels. The feature vectors are obtained using the supervised method with pseudo-labels that contribute to an ambiguous atmosphere. Further, clustering is performed based on these feature vectors. Experimental analyses show that our method outperforms conventional methods in human-like clustering on datasets manually classified by humans.
\end{abstract}

\section{Introduction}
\label{sec:first}
The use of illustrations has increased in advertisements and social media platforms, such as Twitter and Pixiv. However, most tags describe only the content of the illustration, with few mentioning its atmosphere. Expressing the atmosphere of an illustration in words is challenging; moreover, direct description by adding tags can be difficult, and finding illustrations in the desired atmosphere can be time-consuming. 
\par{}Conventional image retrieval methods that are based on textual information \cite{cbmir}, \cite{cbir}, or image features \cite{cnnir} \cite{medical1}, \cite{medical2}  \cite{imagenet} are widely known. Many studies have achieved high accuracy in image classification by training Convolutional Neural Networks (CNN) on unlabeled image datasets \cite{unsupervised1}, \cite{unsupervised2}. However, describing the atmosphere in words is not straightforward; moreover, its determination directly from image content is also challenging. 

Contrary to the supervised classification, clustering methods can divide an image set into clusters without supervision. 
Most of the clustering methods cluster images with similar features without definite classes by using feature vectors obtained from CNNs or handcrafted features \cite{clustering}.  However, there is no guarantee that images with similar atmospheres can be clustered, and there is a possibility that grouping will be performed against users' intentions.

\par{}Considering the above, this paper addresses a task where it is difficult to label images directly.
We propose an atmosphere-based clustering method that combines supervised and unsupervised learning, which discriminates images with similar atmospheres by assigning multiple pseudo-labels that indirectly contribute to the atmosphere of the images and training them using a CNN. The output of the CNN is used as a feature vector, and the images are clustered using the k-means method.
The experimental results are evaluated by measuring, using entropy, the similarity of the clustering results and the human-perceived atmosphere-based clustering. The proposed method demonstrates higher performance in clustering illustrations than conventional methods.

\section{Proposed Method}
\label{sec:proposedmethod}
\begin{figure}[t]
\centering
\includegraphics[clip,width=0.45\textwidth]{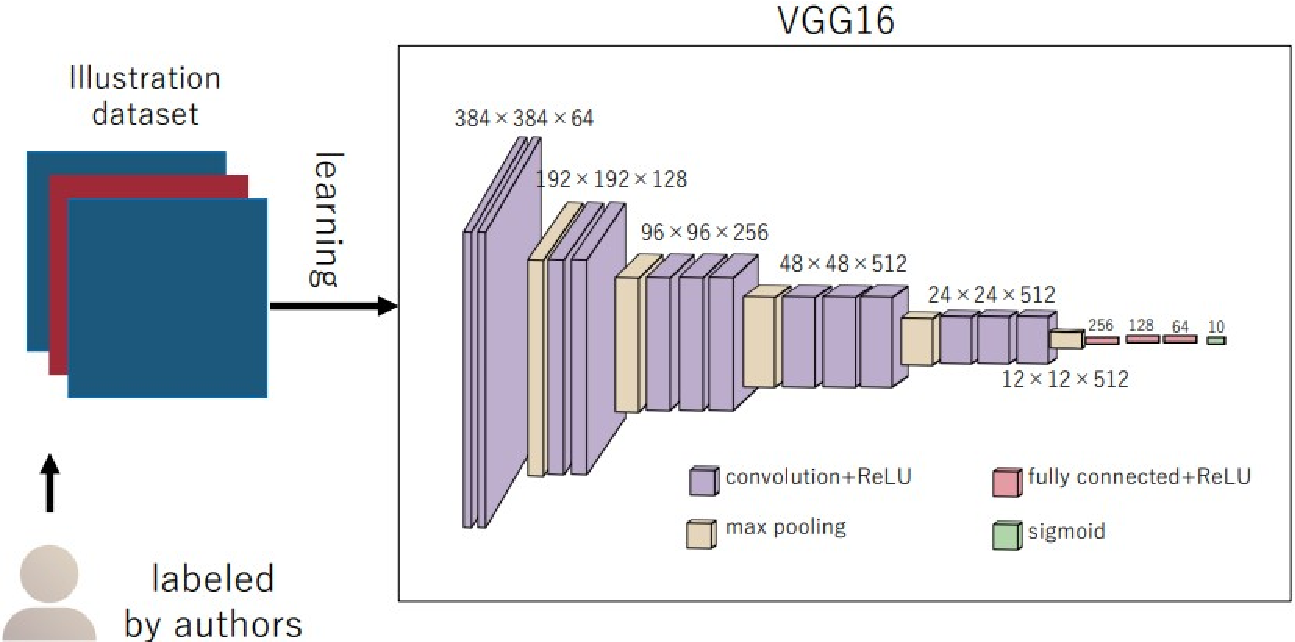}
\caption{An approach to model building to predict illustration pseudo-labels.}
\label{fig:1}
\end{figure}

\begin{figure}[t]
\centering
\includegraphics[clip,width=0.4\textwidth]{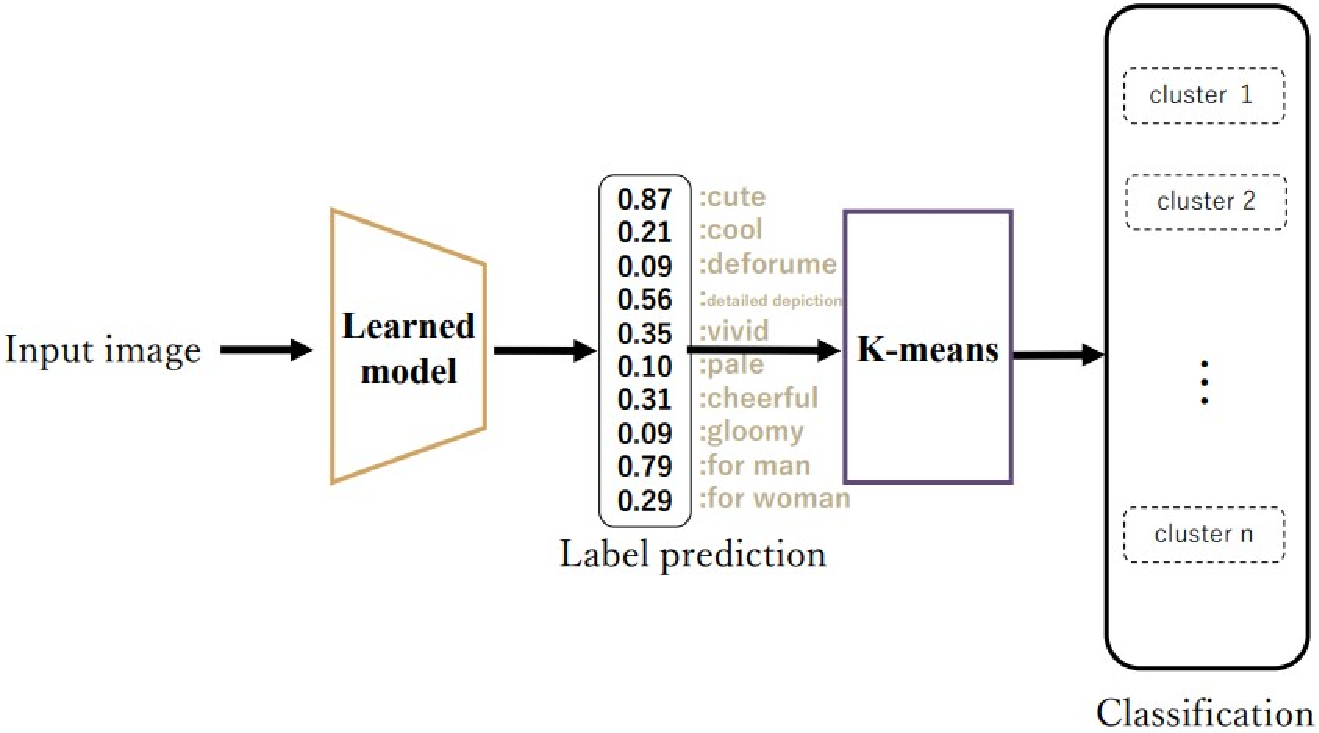}
\caption{Overview of a method for classifying input images by similar atmosphere.}
\label{fig:2}
\end{figure}

It is not easy to verbalize a vague concept such as the 'atmosphere' of an illustration in short words, and hence it is not straightforward to give it a clear label. To solve this problem, instead of conventional image classification using neural networks, which predicts a correct label, the proposed method first predicts pseudo-labels, and the prediction is used as a feature vector (Fig. \ref{fig:1}). Then, the feature vectors are used for clustering (Fig. \ref{fig:2}) to decrease the distance between images with similar atmospheres.
The outline of the proposed method is illustrated in Figures \ref{fig:1} and \ref{fig:2}. Details of each step are given below.

\par{}In the first step, instead of directly assigning labels to images, we assign the pseudo-labels that may affect the atmosphere of images; these are not the final class labels and are referred to as pseudo-labels in this study. The examples of the pseudo-labels are shown in Table \ref{table:1} .
Multiple pseudo-labels are assigned to illustrations to represent the atmosphere. Then, a model for predicting the pseudo-labels of the illustrations is constructed, as shown in Fig. \ref{fig:1}, in which the model, the pre-trained VGG16 \cite{vgg} network based on ImageNet, is fine-tuned using the illustrations with the assigned pseudo-labels. The predicted outputs are then used as feature vectors for the next step.

\par{}
The method of the previous step is applied to the training images, and the images are clustered based on the predicted label outputs as feature vectors using the K-means method. Here, the clustering is based on the assumption that the difference in feature vectors directly represents the difference in atmosphere. The number of clusters K is specified in advance. In the inference stage, the label outputs obtained by the CNN in the first step are used as features and classified by simply measuring the distance from the center of gravity of the K classes (Fig. \ref{fig:2}).

\begin{table*}[t]
\caption{Examples of pseudo-labels}
\label{table:1}
\begin{center}
\begin{tabular}{lccccccc}
\hline
Dataset&\multicolumn{7}{c}{Label name}\\
\hline\noalign{\vskip.5mm}\hline
Twitter&
cute&lovely&cool&beautiful&deformed&detailed dipiction&cheerful\\
&gloomy&for man&for woman&happy&fellow&love\\
\hline
Danbooru&
angry&casual&chibi&depth of field&highres&everyone&lowres\\
&embarrassed&scenery&simple background&sketch&smile&solo&sparkle\\
\hline
\end{tabular}
\end{center}
\end{table*}

\section{Dataset Construction}
\label{sec:dataset}
Illustrations were collected from Twitter and Danbooru websites\footnote{https://danbooru.donmai.us}. Two datasets were created by assigning each illustration multiple pseudo-labels representing their respective atmosphere.

\subsection{Dataset created using Twitter}
\label{sec:twitter}
Images posted from Twitter with the Japanese tag meaning "I will put up the most stretched picture of the year" from 2021/10/24 to 2021/11/02 and with more than 100 "likes" were collected. Images, such as photographs and similar images from the same contributor, were removed from the dataset. After that, 13 pseudo-labels that contributed to the identification of the atmosphere of the illustration were created and assigned to each illustration collected using the procedure mentioned above. Assigning more than one label was possible in this case. 
Furthermore, images that had not been assigned a single pseudo-label were deleted. Therefore, a dataset consisting of 673 illustrations with pseudo-labels was created.

\subsection{Dataset created using Danbooru}
\label{sec:danbooru}
This study selected labels contributing to the identification of atmosphere among the labels assigned to more than one million images in Danbooru. Furthermore, the images were indiscriminately collected from the illustrations to which one or more of the selected labels had been assigned. Finally, 18 selected pseudo-labels were assigned, and a dataset consisting of 966 illustrations was created.

\subsection{Expansion of the training data}
\label{sec:augmentation}
As a preprocessing step, Images with aspect ratios much farther apart than 1:1 were removed from the dataset because the posted illustrations were unevenly sized. All other images were resized to 384 x 384 pixels by resizing and cropping.
This study used more than a dozen labels as pseudo-labels to be assigned to the dataset; however, the number of images assigned to each label was different. Additionally, the dataset did not have sufficient data, resulting in an imbalance. Therefore, to alleviate the imbalance, data augmentation methods were used for the training data as follows.
\par{}First, the geometric transformation was applied to the collected illustrations. Nine new images were generated from one image using rotations between -45 ° and 45 °, vertical translations, and horizontal flips. Furthermore, the Multi-label Synthetic Minority Over-sampling Technique (MLSMOTE) \cite{mlsmote}, was used to generate 284 and 320 new data for the Twitter and the Danbooru datasets, respectively. MLSMOTE is an extension of the Synthetic Minority Over-sampling Technique (SMOTE) \cite{smote}.

\section{Experiment}
\label{sec:experiment}

\subsection{Summary}
\label{sec:outline}
First, the dataset was divided into training, validation, and test data in the ratio 5:3:2. Next, the model described in Fig. \ref{fig:1} was trained. The number of epochs, batch size, and learning rate was set to 50, 64, and 0.0001, respectively. We used  Adam \cite{adam} and binary cross-entropy loss as the optimizer and loss function, respectively.
\par{}The trained model was used to output the feature vectors of the test data. The number of feature vector outputs is equal to the number of pseudo-labels that can be assigned to the illustration. Each feature vector represents the respective label strength of the illustration.
The output results were then input into the k-means method for clustering. Each cluster in the clustering result was a set of illustrations that the proposed method evaluated to have a similar atmosphere. In this experiment, the number of clusters was set to four or five.

\subsection{Evaluation Method}
\label{sec:evaluation}
To the best of our knowledge, there is no conventional method for classifying images based on atmosphere. Therefore, two comparative experiments are conducted here to examine the validity of the feature vectors obtained by the CNN in the first step. 

In the first experiment, the feature vectors are replaced with conventional feature vectors for comparison. In conventional approaches, the intermediate output of pre-trained CNNs such as VGG is generally used as the image feature, but it is unsuitable for our clustering task because its dimensions are too high. Therefore, we use BoVW, one of the most popular methods for image recognition.
Specifically, the feature vectors of illustrations are extracted by BoVW and used for clustering using the k-means method, and then the results are compared (comparative method 1 in Sec.\ref{sec:result}).
The second experiment examines the performance difference between the clustering results with multi-hot correct labels as input (comparative method 2 in Sec.\ref{sec:result23}, respectively).

The performance of the proposed method is evaluated using the silhouette coefficient, which evaluates the clustering performance. Furthermore, it is evaluated using the
entropy, which indicates the degree of variation of the obtained clusters with respect to correct data. 
A description of each criterion is given below.

\subsubsection{Silhouette coefficient}
\label{sec:silhouette}
The silhouette coefficient is a measure of the clustering performance. It takes values in the range [-1,1], with 1 being the best value. The closer the data belong to the same cluster and the farther the data belong to different clusters, the better the silhouette coefficient.

\subsubsection{Entropy}
\label{sec:entropy}
In this study, entropy measured the degree of variability in clustering results relative to the correct data. Entropy was normalized to [0,1] (the base of the logarithm was normalized to $S$ when calculating entropy).
The entropy is calculated as follows:
First, the probability that an image in the proposed method's $j$-th cluster $B_j$ is contained in the $i$-th cluster $A_i$ of the correct data is determined by 
$ P(E_{i,j})=n(B_j\cap A_i) / n(B_j)$, where $n(\cdot)$ represents the number of images in a cluster. 
Next, based on the probability, we calculate the entropy $H_j$ of the $j$-th cluster $B_j$ in terms of how much it varies with respect to the correct data. Let $S$ be the number of clusters in the subjective classification, $H_j$ is calculated by 
$H_j=\sum_{i=1}^{S}P(E_{i,j})\log_SP(E_{i,j})$. 
Finally, the averaged entropy value $H$ is obtained by multiplying the probability that an image in the test data is included in a cluster $B_j$ of the proposed method: $H=\sum_{j=1}^{K}H_j\frac{n(B_j)}{N}$, where $N$ is the total number of images in the test data, and $K$ is the number of clusters in the method.

\subsubsection{Correct data for entropy evaluation}
To evaluate the methods with entropy, correct data grouped based on similarity of atmosphere are needed.
Two types of data were created as the correct data to evaluate the closeness of the clustering of illustrations by the proposed method to human perception. 

First, the authors manually classified the test data by images with similar atmospheres. 
Second, eight raters not involved in the study then classified the images to produce the second set of correct data. This type of correct data was prepared by asking the raters to classify the images under two separate conditions. The first is classification with the condition that the number of images in each classified group must be three or more (Condition 1), and the second is a classification with the condition that the number of groups must be eight (Condition 2). Based on these conditions, the illustrations were classified into groups that were considered similar in atmosphere. This classification was performed only on the Twitter dataset.

\subsection{Experiment 1}
\label{sec:result}
The clustering performance of the proposed method and the comparison method was evaluated based on the silhouette coefficient. Table \ref{table:2} presents the results. The results of the silhouette coefficient evaluation demonstrated that the proposed method performed better than the comparison method in all cases of the two datasets.

Table \ref{table:3} and  \ref{table:4} show the entropy evaluation when the data were used as the correct response for the two types of datasets. The results showed that the dataset created from Twitter had a higher entropy value than that of Danbooru. Moreover, the difference between the entropy values of the proposed method and the comparison method was larger for the dataset created from Twitter than for that of Danbooru, suggesting that illustrations collected from Twitter were more challenging in evaluating the atmosphere than those from Danbooru. In other words, the more complex illustrations, the more human-like clustering of the proposed method was than conventional methods.

\begin{table}[h]
        \caption{Silhouette coefficient for each clustering result.}
        \label{table:2}
    \begin{center}
    \begin{tabular}{llcc}
   \hline
    method&4 clusters&5 clusters\\
    \hline\noalign{\vskip.5mm}\hline
    proposed method (Twitter)&0.35&0.32\\
    comparative method 1 (Twitter)&0.12&0.12\\
    \hline
    proposed method (Danbooru)&0.39&0.40\\
    comparative method 1 (Danbooru)&0.11&0.08\\
    \hline
    \end{tabular}
    \end{center}
\end{table}
\begin{table}[h]
	\caption{Entropy with classification by author as correct data}
	\label{table:3}
    \begin{center}
    \begin{tabular}{lcc}
    \hline
    method&4 clusters&5 clusters\\
    \hline\noalign{\vskip.5mm}\hline
    proposed method (Twitter)&0.79&0.78\\
    comparative method 1 (Twitter)&0.89&0.86\\
    \hline
    proposed method (Danbooru)&0.70&0.68\\
    comparative method 1 (Danbooru)&0.74&0.71\\
    \hline
    \end{tabular}
    \end{center}
\end{table}
\begin{table}[h]
	\caption{Entropy with objective evaluation as the data for correct answers (Twitter dataset). }
	\label{table:4}
    \begin{center}
    \begin{tabular}{lcc}
    \hline
    method&4 clusters&5 clusters\\
    \hline\noalign{\vskip.5mm}\hline
    proposed method (cond. 1)&0.79&0.77\\
    comparative method 1 (cond. 1)&0.84&0.82\\
    \hline
    proposed method (cond. 2)&0.79&0.77\\
    comparative method 1 (cond. 2)&0.85&0.82\\
    \hline
    \end{tabular}
    \end{center}
\end{table}

\subsection{Experiments 2}
\label{sec:result23}
Next, to further verify the validity of the features in this study, numerical experiment was conducted using the correct labels, in which the correct labels are directly input as features to the K-means method (comparative method 2).  For the experiments, we used  the correct data generated from the classification by eight people.

Table\ref{table:6} shows a comparison with a method in which the correct answer labels are input directly into K-means. Naturally, the results of the proposed method are slightly lower than those of the K-means method. The results are slightly lower than those of the proposed method, but they show that the proposed method is able to cluster images without pseudo-labels to similar moods with almost the same accuracy as when pseudo-labels are added.

\begin{table}[h]
	\caption{Entropy when the correct labels are entered into the K-means method as the comparison method. }
	\label{table:6}
    \begin{center}
    \begin{tabular}{lcccc}
    \hline
    method&\multicolumn{2}{c}{condition 1}&\multicolumn{2}{c}{condition 2}\\
        &\multicolumn{2}{c}{\#cluster}&\multicolumn{2}{c}{\#cluster}\\
    &4 &5 &4 &5 \\
    \hline\noalign{\vskip.5mm}\hline
    proposed method&0.79&0.77&0.79&0.77\\
    comparative method 2&0.78&0.77&0.78&0.76\\
    \hline
    \end{tabular}
    \end{center}
\end{table}

\section{Conclusions}
\label{sec:conclusion}
This study created two new datasets with multiple pseudo-labels contributing to the atmosphere to classify illustrations by similar atmospheres in a human-like state. We proposed a clustering method using a CNN and the k-means method. The proposed method uses the label intensities output by the CNN as input and clusters each image based on the closest distance between the feature vectors. The clustering performance was evaluated using silhouette coefficients. Additionally, we evaluated by entropy the similarity of the clustering results of the proposed method to the clustering of illustrations based on the human perception of the atmosphere, using the correct answer data created based on the evaluations of several people.

\par{}In the future, we will consider extending the dataset to suggest illustrations that are close to the desired atmosphere, create a new dataset based on the opinions of more people using crowdsourcing, and use the dataset to recommend illustrations for individual users.

\section*{Acknowledgement}
This Work was supported by MEXT Promotion of Distinctive Joint Research Center Program Grant Number\#
JPMXP 0621467946.

\vfill\pagebreak

\bibliographystyle{IEEEbib}
\bibliography{strings,refs}

\begin{thebibliography}{10}

\bibitem{cbmir}
A.~Kumar, J.~Kim, W.~Cai, M.~J. Fulham, and D.~Feng,
\newblock ``Content-based medical image retrieval: {A} survey of applications
  to multidimensional and multimodality data,''
\newblock {\em J. Digit. Imaging}, vol. 26, no. 6, pp. 1025--1039, 2013.

\bibitem{cbir}
X.~Li, J.~Yang, and J.~Ma,
\newblock ``Recent developments of content-based image retrieval {(CBIR)},''
\newblock {\em Neurocomputing}, vol. 452, pp. 675--689, 2021.

\bibitem{cnnir}
F.~Radenovic, G.~Tolias, and O.~Chum,
\newblock ``Fine-tuning {CNN} image retrieval with no human annotation,''
\newblock {\em {IEEE} Trans. Pattern Anal. Mach. Intell.}, vol. 41, no. 7, pp.
  1655--1668, 2019.

\bibitem{medical1}
H.~Xie, H.~Shan, W.~Cong, X.~Zhang, S.~Liu, R.~Ning, and G.~Wang,
\newblock ``Dual network architecture for few-view {CT} - trained on imagenet
  data and transferred for medical imaging,''
\newblock {\em CoRR}, vol. abs/1907.01262, 2019.

\bibitem{medical2}
M.~Graziani, V.~Andrearczyk, and H.~Muller,
\newblock ``Visualizing and interpreting feature reuse of pretrained cnns for
  histopathology,''
\newblock in {\em {IMVIP} 2019: Irish Machine Vision and Image Processing
  Conference Proceedings}. 2019, pp. 231--234, Dublin, Ireland, Irish Pattern
  Recognition and Classification Society.

\bibitem{imagenet}
J.~Deng, W.~Dong, R.~Socher, L.~Li, K.~Li, and L.~Fei{-}Fei,
\newblock ``Imagenet: {A} large-scale hierarchical image database,''
\newblock in {\em 2009 {IEEE} Computer Society Conference on Computer Vision
  and Pattern Recognition {(CVPR} 2009), 20-25 June 2009, Miami, Florida,
  {USA}}. 2009, pp. 248--255, {IEEE} Computer Society.

\bibitem{unsupervised1}
X.~Ji, A.~Vedaldi, and J.~F. Henriques,
\newblock ``Invariant information clustering for unsupervised image
  classification and segmentation,''
\newblock in {\em 2019 {IEEE/CVF} International Conference on Computer Vision,
  {ICCV} 2019, Seoul, Korea (South), October 27 - November 2, 2019}. 2019, pp.
  9864--9873, {IEEE}.

\bibitem{unsupervised2}
C.~Doersch, A.~Gupta, and A.~A. Efros,
\newblock ``Unsupervised visual representation learning by context
  prediction,''
\newblock in {\em 2015 {IEEE} International Conference on Computer Vision,
  {ICCV} 2015, Santiago, Chile, December 7-13, 2015}. 2015, pp. 1422--1430,
  {IEEE} Computer Society.

\bibitem{clustering}
Amit Saxena, Mukesh Prasad, Akshansh Gupta, Neha Bharill, Om~Prakash Patel,
  Aruna Tiwari, Meng~Joo Er, Weiping Ding, and Chin-Teng Lin,
\newblock ``A review of clustering techniques and developments,''
\newblock {\em Neurocomputing}, vol. 267, pp. 664--681, 2017.

\bibitem{vgg}
K.~Simonyan and A.~Zisserman,
\newblock ``Very deep convolutional networks for large-scale image
  recognition,''
\newblock in {\em 3rd International Conference on Learning Representations,
  {ICLR} 2015, San Diego, CA, USA, May 7-9, 2015, Conference Track
  Proceedings}, Yoshua Bengio and Yann LeCun, Eds., 2015.

\bibitem{mlsmote}
F.~Charte, A.~J. Rivera, M.~J. del Jesus, and F.~Herrera,
\newblock ``{MLSMOTE:} approaching imbalanced multilabel learning through
  synthetic instance generation,''
\newblock {\em Knowl. Based Syst.}, vol. 89, pp. 385--397, 2015.

\bibitem{smote}
N.~V. Chawla, K.~W. Bowyer, L.~O. Hall, and W.~P. Kegelmeyer,
\newblock ``{SMOTE:} synthetic minority over-sampling technique,''
\newblock {\em J. Artif. Intell. Res.}, vol. 16, pp. 321--357, 2002.

\bibitem{adam}
D.~P. Kingma and J.~Ba,
\newblock ``Adam: {A} method for stochastic optimization,''
\newblock in {\em 3rd International Conference on Learning Representations,
  {ICLR} 2015, San Diego, CA, USA, May 7-9, 2015, Conference Track
  Proceedings}, Yoshua Bengio and Yann LeCun, Eds., 2015.

\end{thebibliography}

\end{document}